\documentclass{article}

\usepackage{arxiv}

\usepackage[utf8]{inputenc} 
\usepackage[T1]{fontenc}    
\usepackage{hyperref}       
\usepackage{url}            
\usepackage{booktabs}       
\usepackage{amsfonts}       
\usepackage{nicefrac}       
\usepackage{microtype}      
\usepackage{lipsum}
\usepackage{graphicx}
\usepackage{multirow}
\usepackage{threeparttable}
\graphicspath{ {./images/} }

\title{Explainable Rule Application via Structured Prompting: A Neural-Symbolic Approach\thanks{Accepted for publication at the 29th International Conference on Knowledge-Based and Intelligent Information \& Engineering Systems (KES 2025).}}

\author{
 Albert Sadowski \\
  Faculty of Electronics and Information Technology\\
  Warsaw University of Technology, Poland\\
  Warsaw, Poland \\
  \texttt{albert.sadowski.stud@pw.edu.pl} \\
   \And
 Jarosław A. Chudziak \\
  Faculty of Electronics and Information Technology\\
  Warsaw University of Technology\\
  Warsaw, Poland \\
  \texttt{jaroslaw.chudziak@pw.edu.pl} \\
}

\begin{document}
\maketitle
\begin{abstract}
Large Language Models (LLMs) excel in complex reasoning tasks but struggle with consistent rule application, exception handling, and explainability, particularly in domains like legal analysis that require both natural language understanding and precise logical inference. This paper introduces a structured prompting framework that decomposes reasoning into three verifiable steps: entity identification, property extraction, and symbolic rule application. By integrating neural and symbolic approaches, our method leverages LLMs’ interpretive flexibility while ensuring logical consistency through formal verification. The framework externalizes task definitions, enabling domain experts to refine logical structures without altering the architecture. Evaluated on the LegalBench hearsay determination task, our approach significantly outperformed baselines, with OpenAI \textit{o}-family models showing substantial improvements - \textit{o1} achieving an F1 score of 0.929 and \textit{o3-mini} reaching 0.867 using structured decomposition with complementary predicates, compared to their few-shot baselines of 0.714 and 0.74 respectively. This hybrid neural-symbolic system offers a promising pathway for transparent and consistent rule-based reasoning, suggesting potential for explainable AI applications in structured legal reasoning tasks.
\end{abstract}


\section{Introduction}

Large Language Models (LLMs) have demonstrated impressive capabilities in complex reasoning tasks \cite{Wei2022} \cite{Bubeck2023}, showing potential to transform domains requiring sophisticated text analysis and inference. Their ability to process natural language inputs and generate coherent responses makes them seemingly ideal for tasks that traditionally demand human-level reasoning (e.g., legal analysis \cite{Blair-Stanek2025}, regulatory compliance, and evidence-based decision making). However, despite their power, LLMs face significant limitations when applications require consistent rule application \cite{Mu2023} \cite{GSMSymbolic}, transparent reasoning processes, and explainable outcomes.

The fundamental challenge lies in the tension between neural and symbolic approaches to reasoning \cite{Wei2025} \cite{Lalwani2024}. Neural methods like LLMs offer flexibility and generalization across diverse inputs but often struggle with consistency and verifiability \cite{Bubeck2023}. In contrast, symbolic systems provide formal guarantees but traditionally require structured inputs and lack adaptability to natural language \cite{Calanzone2024}. This dichotomy becomes particularly apparent in rule application scenarios where both precise logical inference and natural language understanding are essential \cite{BenchCapon1997}.

Three specific challenges emerge in deploying LLMs for rule-based reasoning tasks \cite{Kant2025} \cite{Borazjanizadeh2024}. First, while LLMs can recognize patterns in training data, they lack consistent mechanisms to apply rules with logical rigor across novel situations \cite{Mu2023}. Second, exception handling - determining when general rules don't apply due to special conditions-presents significant difficulties \cite{DiSorbo2025} as it requires both rule comprehension and contextual awareness. Third, the black-box nature of neural architectures makes reasoning paths opaque \cite{Dahl2024}, limiting human oversight and preventing verification of the logical steps leading to a conclusion.

We propose a novel structured prompting framework that bridges these neural and symbolic paradigms. By decomposing complex reasoning into three discrete steps-entity identification, property extraction, and rule application through symbolic verification-our approach leverages the complementary strengths of both methodologies. LLMs handle the interpretation of natural language inputs and extraction of relevant entities and properties, while symbolic verification ensures logical consistency in rule application. This separation of concerns creates a system that maintains interpretative flexibility through externalized predicate definitions while providing structured reasoning paths that enable human oversight.

Our contributions include a neural-symbolic framework implementing structured decomposition principles that separates natural language processing (handled by LLMs) from logical verification (handled by SMT solvers) across three distinct steps of entity identification, property extraction, and rule application. We externalize task definitions with complementary predicates, enabling domain experts to refine logical structures without architectural changes, and provide experimental validation demonstrating significant performance improvements (12-14\% pp. F1 improvement on OpenAI \textit{o}-family models). Additionally, we analyze how different model architectures respond to structured prompting and examine the precision-recall tradeoffs.

The remainder of this paper is organized as follows: Section \ref{section:background} reviews the background and related work in LLM reasoning, prompting techniques, and neural-symbolic integration. Section \ref{section:framework} illustrates the framework with a case study on hearsay determination (determining whether evidence constitutes inadmissible out-of-court statements) in legal reasoning. Section \ref{section:experiment} describes our experimental design, while Section \ref{section:results} presents the results and analysis. We discuss limitations and broader implications in Section \ref{section:discussion} before concluding in Section \ref{section:conclusion}.

\section{Background and Related Work}
\label{section:background}

LLMs have demonstrated impressive capabilities in complex reasoning tasks \cite{Wei2022}, yet face significant challenges with logical inference, consistent rule application, and exception handling \cite{Bubeck2023}. While reasoning abilities emerge as model size increases, they lack transparency and verifiability-particularly problematic in domains requiring explanation and justification.

In legal applications, LLMs show promise but mixed results. On LegalBench \cite{LegalBench}, state-of-the-art models like GPT-4 demonstrate capabilities across various legal domains but lack the consistency and explainability required for real-world use. As Kant et al. (2024) note, legal contexts demand ``higher accuracy, repeatability, and transparency'' \cite{Kant2025}. The Mata v. Avianca case-where an attorney unknowingly submitted ChatGPT-generated fictitious judicial opinions-underscores these risks \cite{Aidid2024}.

Key challenges include the models' auto-regressive nature leading to ``greedy'' word-by-word generation rather than planning \cite{Borazjanizadeh2024} and legal hallucinations such as generating nonexistent citations and misinterpreting provisions \cite{Magesh2024}. Recent approaches combining LLMs with domain-specific knowledge retrieval \cite{Matak2025} show promise but explainability and consistent rule application remain significant obstacles.

\subsection{Prompting Techniques for Complex Reasoning}

To improve LLMs' reasoning capabilities, researchers have developed various prompting techniques. Chain-of-Thought (CoT) prompting \cite{Wei2022} decomposes complex reasoning into intermediate steps, significantly improving performance on tasks requiring multi-step inference. By providing a few demonstrations of step-by-step reasoning, CoT enables models to mimic this process on new problems, often leading to dramatic performance improvements, especially in larger models.

Beyond CoT, other specialized prompting techniques have emerged. Program-Aided Language Models (PAL) \cite{Gao2022} leverages LLMs to generate executable code that solves reasoning problems, combining the flexibility of natural language with the precision of programming languages. ReAct \cite{Yao2022} interleaves reasoning and acting, allowing models to interact with external environments while building coherent reasoning traces.

In the legal domain specifically, Servantez et al. (2024) introduced the Chain of Logic prompting method, which ``decomposes legal reasoning into independent logical steps and recomposes them to form coherent conclusions for rule-based legal evaluation'' \cite{Servantez2024}. Similarly, the InsurLE framework by Cummins et al. (2025) uses controlled natural language to codify insurance contracts while preserving key syntactic nuances and exposing the underlying formal logic \cite{Cummins2025}.

\subsection{Neural-Symbolic Integration Approaches}

Neural-symbolic approaches combine the learning capabilities of neural networks with the reasoning power of symbolic systems. For legal applications, this integration is particularly promising as it combines LLMs' natural language understanding with the formal reasoning capabilities of logic-based systems.

Tan et al. (2024) enhanced LLM reasoning through a self-driven Prolog-based CoT mechanism that iteratively refines logical inferences in legal tasks \cite{Tan2024}. Similarly, Wei et al. (2025) proposed a hybrid neural-symbolic framework that synergizes neural representations with explicit logical rules to improve legal reasoning in automated systems \cite{Wei2025}. Similar integration approaches have been explored in multi-agent contexts, where Answer Set Programming combined with graph knowledge bases enables enhanced collaborative reasoning capabilities \cite{Kostka2024}.

The work by Lalwani et al. (2024) on NL2FOL demonstrates a structured, step-by-step pipeline to translate natural language inputs into first-order logic representations using LLMs at each step. Their approach addresses key challenges in this translation process, including integrating implicit background knowledge. By leveraging structured representations, they use Satisfiability Modulo Theory (SMT) solvers to reason about the logical validity of natural language statements \cite{Lalwani2024}.

Alonso and Chatzianastastiou (2024) demonstrated that ``embedding logical rules into neural frameworks can enhance the interpretability and robustness of legal text analysis'' \cite{Noguer2024}. Calanzone et al. (2024) developed an integration approach that enforces logical consistency by incorporating external constraint sets into LLM outputs \cite{Calanzone2024}.

\begin{table*}[h]
\caption{Comparison of neural-symbolic methods along domain, decomposition, and symbolic reasoning dimensions.}
\begin{tabular*}{\hsize}{@{\extracolsep{\fill}}lp{2.5cm}p{4cm}p{4cm}@{}}
\toprule
\textbf{Method} & \textbf{Domain} & \textbf{Decomposition Strategy} & \textbf{Symbolic Formalism} \\
\midrule
Thought-Like-Pro~\cite{Tan2024} & General & Explicit multi-step (CoT-like) & Prolog-based inference \\
Wei et al. (2025)~\cite{Wei2025} & Legal (civil) & Structured pipeline & FOL with custom legal rules \\
NL2FOL~\cite{Lalwani2024} & General & Stepwise formalization & FOL + SMT solver \\
Alonso \&\\ Chatzianastasiou (2024))~\cite{Noguer2024} & Legal (contracts) & Logic-LLM integration & Logic rule templates \\
Calanzone et al. (2024)~\cite{Calanzone2024} & General & End-to-end constraint loss & Logical consistency constraints \\
\bottomrule
\end{tabular*}
\label{tab:ns_comparison}
\end{table*}

As these studies illustrate, neural-symbolic systems vary widely in how they structure reasoning tasks and integrate symbolic logic. Table~\ref{tab:ns_comparison} provides a comparative overview of representative approaches, highlighting key differences in domain focus, decomposition strategy, and symbolic formalism.

\subsection{Legal Rule Application Challenges}

Legal rule application presents unique challenges that test the limits of current AI systems. Legal rules contain inherent complexity through their ``open texture'' \cite{BenchCapon1997} and numerous exceptions, with legal concepts often having deliberately vague boundaries requiring contextual interpretation. Exceptions and exclusions (like those in hearsay evidence) demand nuanced understanding of both the rule and its application context \cite{Blair-Stanek2025}. Additionally, legal reasoning frequently requires counterfactual reasoning - determining what would have happened under different circumstances - a form of analysis that remains particularly challenging for current AI systems \cite{Hu2025}. This involves hypothetical scenario evaluation that goes beyond pattern recognition to causal understanding.

Effective legal reasoning also necessitates integrating diverse knowledge sources (statutes, case law, regulations) \cite{Dahl2024} while providing transparent explanations for conclusions. Legal applications face ``persistent transparency, ethical compliance, and reliability challenges'' \cite{Dahl2024}, as decisions must be auditable and justifiable to various stakeholders. The proposed structured prompting approach addresses these challenges by breaking legal reasoning into discrete, manageable steps while maintaining interpretability through externalized predicate definitions, potentially offering a path toward more reliable and explainable legal AI systems.

\section{Structured Decomposition Framework}
\label{section:framework}

Decomposition-based approaches have long been recognized in cognitive science and AI as effective strategies for complex problem solving \cite{Newell1972, Sacerdoti1974}. Building on this foundation, our structured decomposition framework applies these well-established principles to address a fundamental challenge in AI reasoning: while LLMs have demonstrated impressive capabilities in handling natural language inputs, they often struggle with consistent rule application \cite{GSMSymbolic, Mu2023}.

The framework bridges neural and symbolic approaches by leveraging their complementary strengths-neural components (LLMs) excel at pattern recognition and extracting information from unstructured text, while symbolic components provide formal verification mechanisms ensuring logical consistency and transparency. By separating term identification and predicate extraction (handled by LLMs) from rule application (handled through symbolic verification), we create a system that maintains flexibility in interpretation while enforcing logical consistency in rule application.

While our evaluation focuses on legal reasoning, the structured decomposition architecture itself is domain-agnostic and applicable to any context requiring rule-based inference. As shown in Table~\ref{tab:ns_comparison}, existing neural-symbolic systems differ in their modularity and the depth of symbolic integration. Our approach emphasizes a clear separation of concerns between language interpretation and logical verification, resulting in a transparent and adaptable framework that contrasts with more tightly coupled or domain-specific designs.

\subsection{Neural-Symbolic Three-Step Architecture}

Building on established decomposition approaches \cite{DecomposedPrompting, LM2}, any reasoning task in our framework begins with a task definition. Domain experts define the terms (i.e., entities to be identified in text), predicates (i.e., relationships or attributes of these entities), and task predicate (i.e., logical rule that determine when specific conditions are met). This task definition establishes the structure for the reasoning process that follows. Terms represent key entities (e.g., people, objects, events), while predicates capture relationships between these entities.

\begin{figure}[h]\vspace*{4pt}
\centerline{\includegraphics[width=\linewidth]{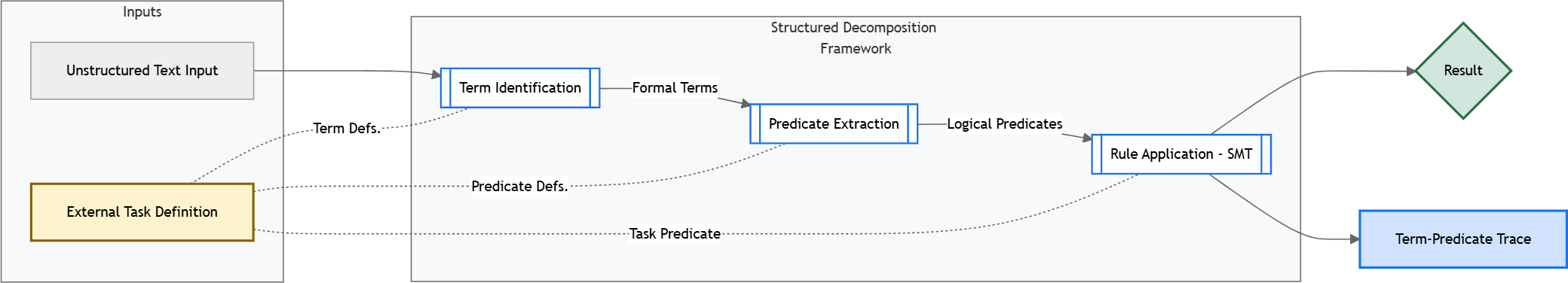}}
\caption{Structured Decomposition Framework. The three-step process flows from Term Identification through Predicate Extraction to Rule Application via SMT solving. Yellow box represents domain expert input (task definition, i.e., terms, predicates, rules), blue boxes show automated LLM/solver processing, and arrows indicate data flow between components.}
\label{fig:structured_decomposition_framework}
\end{figure}

Our framework then decomposes reasoning into three distinct steps, as illustrated in Figure~\ref{fig:structured_decomposition_framework}.

In the \textbf{term identification} step (identifying relevant text spans), the LLM extracts entities from unstructured text that correspond to terms in the reasoning task defined in the task definition. The model identifies text spans that match predefined term definitions, assigns entities to appropriate terms, and provides explanations for each extraction. This step converts unstructured text into structured data points, creating a foundation for subsequent reasoning steps. The entity extraction prompt provides clear definitions and asks the model to justify its selections, making the process transparent and verifiable.

The \textbf{predicate extraction} step (determining logical relationships) determines which predicates apply to the identified terms. The LLM analyzes relationships between identified terms, determines which predefined predicates apply to which terms, and provides explanations for each predicate attribution. This step captures the relational aspects of reasoning, identifying how terms relate to each other according to the domain rules. The predicate extraction prompt explicitly references the identified terms and asks the model to reason about which predicates apply, again providing justifications for each determination.

The final \textbf{rule application} step applies formal logic to determine whether the task predicate is satisfied based on the extracted predicates. This step converts extracted predicates into logical expressions, uses a SMT solver to verify if the task predicate is satisfied, and returns a comprehensive result that includes the logical determination and supporting evidence.

While this symbolic verification step itself is deterministic, it is important to note that the overall system's reliability still depends on the quality of entity and predicate extraction performed by the LLM in the earlier steps. The LLM's interpretation of terms and predicates based on given definitions introduces uncertainty that propagates through the system. However, the symbolic verification component provides deterministic guarantees for the rule application phase given the extracted predicates as input. This hybrid approach creates a clear separation of concerns: LLMs handle the interpretation of natural language and extraction of structured information, while the symbolic solver focuses on rule application. Section~\ref{sec:hearsay_case_study} illustrates this process with concrete example from our hearsay case study (Tables~\ref{tab:entity_extraction} and~\ref{tab:property_extraction}).

\subsection{Externalized Task Predicate Definitions}

Our approach externalizes task predicate definitions as part of the task definition. When domain experts create a task, they define the problem domain, term definitions (entities to extract), predicate definitions (attributes or relationships), and task predicate definition (logical rule satisfaction conditions).

Critically, designers should include complementary predicates (e.g., both \texttt{IsInCourt} and \texttt{IsOutOfCourt}), forcing LLMs to make explicit choices between opposing options rather than defaulting to one. This design choice addresses a fundamental issue in LLM-based systems: the tendency to exhibit confirmation bias by preferentially extracting positive instances of predicates while neglecting their negation. By requiring explicit consideration of both possibilities, we reduce false positives and increase precision in complex domains where nuance is essential.

The implementation of complementary predicates requires balancing completeness with complexity. While comprehensive predicate coverage reduces ambiguity by forcing explicit binary choices, additional predicates increase prompt complexity, potentially overwhelming models beyond a certain threshold. The decision to implement complementary predicates should be treated as a design choice requiring empirical validation, as the optimal balance between logical completeness and cognitive complexity may vary across different model architectures and reasoning capabilities.

The framework uses first-order logic expressions while addressing its limitations in complex domains by allowing task designers to incorporate concepts like deontic operators at different abstraction levels. This balances formal verifiability with expressive power-domain experts can modify task predicates to accommodate evolving interpretations while maintaining both rigor and flexibility in the reasoning process.

\subsection{Case Study: Hearsay Determination}
\label{sec:hearsay_case_study}

To illustrate our framework, we present a hearsay determination example. \href{https://www.uscourts.gov/rules-policies/current-rules-practice-procedure/federal-rules-evidence}{The Federal Rules of Evidence} define hearsay as an ``out-of-court statement introduced to prove the truth of the matter asserted'' - a rule requiring nuanced understanding of statements, context, and purpose. Domain experts can formalize this rule as: $IsHearsay(s, l)$ $\leftrightarrow$ $\Big(IsStatement(s)$ $\land$ $OutOfCourt(s)$ $\land \exists a$ $\big(HasAssertion(s, a)$ $\land$ $IntroducedFor(s, l)$ $\land$ $ProvesTruth(s, l)\big)\Big)$, where $s$ is a communicative act asserting a fact, $l$ is the disputed matter, and $a$ is the factual claim conveyed.

Consider this example from the LegalBench dataset \cite{LegalBench}: \textit{``On the issue of whether Martin punched James, the fact that Martin smiled and nodded when asked if he did so by an officer on the scene.''} Our framework processes this through its workflow: first, the LLM identifies key entities from the text (the legal issue $l$, statement $s$, and assertion $a$) as shown in Table~\ref{tab:entity_extraction}. Next, it determines which predicates apply to these entities (Table~\ref{tab:property_extraction}), recognizing critical properties like whether the statement occurred out-of-court. Finally, the symbolic verifier evaluates the formal hearsay predicate based on these extracted terms and predicates. This transparent pathway from unstructured text to logical conclusion enables fine-grained analysis at each reasoning step, enhancing both accuracy and explainability.

\begin{table*}[b]
\caption{Term Identification Results}
\begin{tabular*}{\hsize}{@{\extracolsep{\fill}}lp{6cm}p{9cm}@{}}
\toprule
\textbf{Term} & \textbf{Text} & \textbf{Explanation} \\
\midrule
$l$ & the issue of whether Martin punched James & This text directly references the disputed matter by posing it as 'the issue of whether...'. \\
$s$ & Martin smiled and nodded when asked if he did so by an officer on the scene & Nodding and smiling in response to a question can be considered non-verbal communicative conduct that asserts a fact. \\
$a$ & Martin punched James & This is the factual claim implied by Martin's non-verbal acknowledgement; it represents the proposition being asserted. \\
\bottomrule
\end{tabular*}
\label{tab:entity_extraction}
\end{table*}

\begin{table*}[b]
\caption{Predicate Extraction Results}
\begin{tabular*}{\hsize}{@{\extracolsep{\fill}}l p{1cm} p{11cm} @{}}
\toprule
\textbf{Predicate Name} & \textbf{Arguments} & \textbf{Explanation} \\
\midrule
\texttt{IsStatement} & $s$ & Under Federal Rules of Evidence, non-verbal conduct qualifies as a ``statement'' when it is assertive conduct intended to communicate information. Martin's smile and nod constitute assertive conduct because they were intentional responses to a direct question, conveying his acknowledgment. \\
\texttt{OutOfCourt} & $s$ & The conduct occurred at the scene with an officer, not in a courtroom proceeding, satisfying the out-of-court requirement for hearsay analysis. \\
\texttt{HasAssertion} & $s$, $a$ & By smiling and nodding, Martin effectively asserted that he punched James ($a$), establishing the propositional content of the statement. \\
\texttt{IntroducedFor} & $s$, $l$ & The evidence is being introduced specifically to resolve the disputed matter ($l$). This predicate evaluates true when the statement's probative value depends on proving the truth of what was asserted. \\
\texttt{ProvesTruth} & $s$, $l$ & Martin's acknowledgment is used as evidence to establish the truth of the assertion that he punched James, rather than for other purposes like impeachment or demonstrating state of mind. \\
\bottomrule
\end{tabular*}
\begin{tablenotes}
\footnotesize
\item Note: Each predicate is evaluated to determine whether it holds for the specified arguments.
\end{tablenotes}
\label{tab:property_extraction}
\end{table*}

\section{Experimental Setup and Evaluation}
\label{section:experiment}
We selected the Hearsay determination task from LegalBench \cite{LegalBench} as the evaluation benchmark for our structured prompting framework. This task requires rigorous application of formal rules and presents an ideal case for evaluating our approach. The hearsay task requires determining whether evidence constitutes inadmissible hearsay under the Federal Rules of Evidence through multi-step reasoning: identifying whether a statement exists (oral, written, or non-verbal assertive conduct); determining if the statement was made out-of-court; and assessing whether the statement is being introduced to prove the truth of the matter asserted.

We use the standard split provided by LegalBench in the form of 'test' and 'train' files, with examples from 'train' used as few-shot exemplars in our prompts while evaluating performance on the 'test' examples. The hearsay dataset contains 4 training examples used as few-shot exemplars and 95 test examples for evaluation, maintaining consistency with the evaluation protocol established in the LegalBench paper \cite{LegalBench}. We evaluate our framework against two established baselines: standard few-shot prompting (models given examples from the LegalBench 'train' file followed by a new test case) and CoT prompting (instructing models to ``think step by step'' with demonstrations of intermediate reasoning). These baselines allow us to assess whether our structured decomposition framework provides benefits beyond general reasoning techniques and determine if explicit rule-based decomposition offers advantages over more flexible approaches.

\subsection{Implementation Details}

We evaluate our framework across six state-of-the-art language models from three providers: OpenAI (\href{https://platform.openai.com/docs/models/gpt-4o-mini}{\textit{gpt-4o-mini-2024-07-18}}, \href{https://platform.openai.com/docs/models/o1}{\textit{o1-2024-12-17}}, and \href{https://platform.openai.com/docs/models/o3-mini}{\textit{o3-mini-2025-01-31}}), Anthropic (\href{https://docs.anthropic.com/en/docs/about-claude/models/overview}{\textit{claude-3-7-sonnet-20250219}}), and Fireworks AI (\href{https://fireworks.ai/models/fireworks/llama-v3p3-70b-instruct}{\textit{llama-v3p3-70b-instruct}} and \href{https://fireworks.ai/models/fireworks/deepseek-v3}{\textit{deepseek-v3}}). Models use deterministic settings (temperature 0.0 where available), with responses structured via \href{https://github.com/pydantic/pydantic}{Pydantic (v2.x)} models through the \href{https://github.com/langchain-ai/langchain}{LangChain (v0.3.21)} framework.

For structured decomposition, we test two variants: standard definitions with primary predicates only, and enhanced definitions including complementary predicates (e.g., both \textit{IsInCourt} and \textit{IsOutOfCourt}). While using LegalBench's dataset, our task definitions were developed by computer scientists rather than legal experts - a limitation in legal precision but demonstrative of our framework's flexibility for domain expert refinement.

The rule application step implements symbolic verification using \href{https://www.nltk.org/}{NLTK's (v3.9.1)} SMT solver. For the hearsay determination task, the logical rule translates to the following expression (accepted by NLTK's SMT solver): \texttt{IsStatement(s) \& IsOutOfCourt(s) \& exists a (HasAssertion(s, a) \& IntroducedForLegalIssue(s, l) \& ProvesTruthOfAssertion(s, l))}. The solver evaluates this formula using predicates extracted by the LLM to determine final classification.

Complete implementation details, including all prompts used in the three-step framework and the SD-Direct ablation study, are available in our GitHub repository: \url{https://github.com/albsadowski/structured-decomposition}. Our evaluation methodology validates against LLM hallucinations through supervised comparison with LegalBench's ground truth labels, with performance metrics directly capturing any instances of incorrect entity identification or predicate attribution.

\section{Results and Analysis}
\label{section:results}

Table \ref{tab:performance_comparison} presents the results across all models and methods. The most significant finding is the exceptional performance of structured decomposition with complementary predicates using the \textit{o1} and \textit{o3-mini} models, achieving an F1 score of 0.929 and 0.867, respectively - substantially higher than both few-shot prompting (0.714, 0.740) and CoT prompting (0.767, 0.658) with the same models.

\begin{table*}[!b]
\centering
\caption{Performance Comparison Across Prompting Strategies and Models on Hearsay Determination Task}
\label{tab:performance_comparison}
\begin{tabular*}{\hsize}{@{\extracolsep{\fill}}l p{3cm} p{2cm} p{1.2cm} p{1.2cm} p{1.2cm} @{}}
\toprule
\textbf{Model} & \textbf{Prompting Strategy} & \textbf{Accuracy} & \textbf{Precision} & \textbf{Recall} & \textbf{F1} \\
\midrule
\multirow{5}{*}{GPT-4o mini} 
& Few-shot & 0.734 & 0.690 & 0.707 & 0.699 \\
& Chain-of-Thought & 0.713 & 0.625 & 0.854 & 0.722 \\
& SD & 0.649 & 0.557 & 0.951 & 0.703 \\
& SD Complementary & 0.745 & 0.793 & 0.561 & 0.657 \\
& SD Direct (no SMT) & 0.777 & 0.778 & 0.683 & 0.727 \\
\midrule
\multirow{5}{*}{o1} 
& Few-shot & 0.787 & 0.862 & 0.610 & 0.714 \\
& Chain-of-Thought & 0.819 & 0.875 & 0.683 & 0.767 \\
& SD & 0.777 & 0.679 & 0.927 & 0.784 \\
& SD Complementary & \textbf{0.936} & \textbf{0.907} & \textbf{0.951} & \textbf{0.929} \\
& SD Direct (no SMT) & 0.840 & 0.760 & 0.927 & 0.835 \\
\midrule
\multirow{5}{*}{o3-mini} 
& Few-shot & 0.798 & 0.844 & 0.659 & 0.740 \\
& Chain-of-Thought & 0.734 & 0.750 & 0.585 & 0.658 \\
& SD & 0.787 & 0.818 & 0.659 & 0.730 \\
& SD Complementary & \textbf{0.830} & \textbf{0.821} & \textbf{0.780} & \textbf{0.867} \\
& SD Direct (no SMT) & 0.777 & 0.679 & 0.927 & 0.784 \\
\midrule
\multirow{5}{*}{Llama 3.3 70B} 
& Few-shot & 0.777 & 0.763 & 0.707 & 0.734 \\
& Chain-of-Thought & 0.766 & 0.806 & 0.610 & 0.694 \\
& SD & 0.532 & 0.482 & 1.000 & 0.651 \\
& SD Complementary & 0.734 & 0.690 & 0.707 & 0.699 \\
& SD Direct (no SMT) & 0.734 & 0.767 & 0.561 & 0.648 \\
\midrule
\multirow{5}{*}{DeepSeek V3} 
& Few-shot & 0.766 & 0.711 & 0.780 & 0.744 \\
& Chain-of-Thought & 0.777 & 0.833 & 0.610 & 0.704 \\
& SD & 0.691 & 0.600 & 0.878 & 0.713 \\
& SD Complementary & 0.713 & 0.646 & 0.756 & 0.697 \\
& SD Direct (no SMT) & 0.755 & 0.714 & 0.732 & 0.723 \\
\midrule
\multirow{5}{*}{Claude 3.7 Sonnet} 
& Few-shot & 0.798 & 0.739 & 0.829 & 0.782 \\
& Chain-of-Thought & 0.840 & 0.861 & 0.756 & 0.805 \\
& SD & 0.660 & 0.563 & 0.976 & 0.714 \\
& SD Complementary & 0.628 & 0.539 & 1.000 & 0.701 \\
& SD Direct (no SMT) & 0.723 & 0.619 & 0.951 & 0.750 \\
\bottomrule
\end{tabular*}
\begin{tablenotes}
\footnotesize
\item Note: SD = Structured Decomposition. Bold values indicate the best performance for each model. Complementary predicates include both positive and negative forms (e.g., \textit{IsInCourt} and \textit{IsOutOfCourt}). Direct approach combines all steps without symbolic verification.
\end{tablenotes}
\end{table*}

\subsection{Impact of Framework Components and Ablation Studies}

Our experiments revealed significant model-dependent responses to different components of our framework. The most notable finding was the consistent performance improvement observed across OpenAI's \textit{o} model family, with both \textit{o1} and \textit{o3-mini} showing substantial gains when using structured decomposition with complementary predicates.

The inclusion of complementary predicates affected performance differently across models. With OpenAI \textit{o}-family models, this approach led to remarkable improvements, with \textit{o1} showing a 14.5 percentage point F1 gain and \textit{o3-mini} demonstrating a 12.7 percentage point improvement compared to the few-shot baseline, confirming our hypothesis about explicit choices reducing ambiguity.

Interestingly, for some models like \textit{Claude 3.7 Sonnet}, complementary predicates had the opposite effect, decreasing overall accuracy while maintaining high recall. This pattern, combined with mixed results across other models, indicates that complementary predicates' effectiveness depends on balancing logical completeness against prompt complexity. Models with stronger reasoning capabilities appear better equipped to handle the additional cognitive load, while others may reach a complexity threshold where additional predicates become counterproductive.

A clear precision-recall tradeoff emerged across multiple models. Smaller models like \textit{GPT-4o mini} exhibited higher recall but lower precision with standard structured decomposition, and higher precision but lower recall with complementary predicates. This pattern was particularly pronounced with \textit{Llama 3.3 70B}, which achieved perfect recall with standard structured decomposition but at the cost of precision (0.482).

Our ablation study isolates the contribution of symbolic verification by comparing our full framework against SD-Direct, which uses identical predicate definitions but combines all reasoning steps in a single LLM call. SD-Direct achieved competitive performance (0.835 F1 with \textit{o1}, 0.784 with \textit{o3-mini}) but consistently underperformed the full structured approach across all high-compatibility models. The 11.2 percentage point improvement (\textit{o1}: 0.929 vs 0.835) demonstrates that architectural decomposition with symbolic verification provides benefits beyond providing predicate definitions to the model.

These findings highlight the importance of matching prompting techniques to specific model architectures and confirm that our framework's components contribute differently to overall performance, with symbolic verification playing a crucial role.

\subsection{Structural Benefits for Explainability}
While our structured decomposition framework showed significant performance improvements with specific models, its design offers inherent advantages for explainability regardless of performance outcomes. The three-step decomposition approach creates natural inspection points that enhance transparency through documented reasoning paths, explicit predicate justification, and verifiable symbolic evaluation. By separating term identification, predicate extraction, and rule application, the reasoning process is naturally documented with each step producing intermediate outputs (as demonstrated in Tables \ref{tab:entity_extraction} and \ref{tab:property_extraction}) that can be independently examined, unlike end-to-end approaches where reasoning is embedded within a single narrative.

The predicate extraction step requires explicit justifications for each determination, creating a record of the reasoning behind critical decisions, while complementary predicates force consideration of both confirming and disconfirming factors. Finally, the symbolic verification step ensures that rule application follows deterministic logical principles, creating consistency in how extracted predicates are evaluated against task definitions.

\section{Discussion}
\label{section:discussion}

Our structured decomposition framework demonstrates potential for addressing challenges in LLM-based rule application, though with important limitations. The hybrid neural-symbolic design attempts to combine LLMs' interpretative flexibility during entity and predicate extraction with the consistency of symbolic verification during rule application. This addresses a fundamental tension in legal reasoning systems - the need for both language adaptability and strict rule adherence.

The externalization of predicate definitions allows domain experts to refine logical structures without changing the underlying architecture, potentially accommodating the ``open texture'' of legal concepts described by Bench-Capon and Visser \cite{BenchCapon1997}. Our experimental results reveal systematic model sensitivity patterns that practitioners should consider when deploying this framework. Models can be categorized into three empirical performance tiers: high-compatibility models (OpenAI \textit{o}-family) demonstrate substantial improvements with structured decomposition (\textit{o1} reaching 0.929 F1 and \textit{o3-mini} achieving 0.867 F1); moderate-compatibility models (Claude, GPT-4o mini) show mixed results with sensitivity to prompt complexity thresholds; and lower-compatibility models (Llama, DeepSeek) exhibit performance degradation, suggesting cognitive load limitations. This pattern indicates that the framework's effectiveness correlates with models' multi-step reasoning capabilities, though the underlying factors remain opaque in closed-source systems. Practitioners should conduct pilot evaluations to assess compatibility before deployment, as the 12-14 percentage point improvements observed with compatible models may not generalize across all architectures.

Several limitations must be acknowledged. First, the framework currently focuses on binary classification tasks, representing only a small subset of legal reasoning challenges. Second, the framework requires explicit predicate definitions to be established before operation, creating a dependency on upfront task formalization that may not capture the full complexity or nuanced interpretations possible in legal reasoning. Third, this work approaches legal reasoning as a challenging technical problem to demonstrate neural-symbolic integration principles rather than developing a deployment-ready legal AI system, with our predicate definitions created by computer scientists rather than legal practitioners. Any practical application would require extensive legal expert involvement and ethical review before deployment.

The structured decomposition creates natural inspection points for domain experts, as detailed in Section 5.2. Even in cases where performance improvements were modest with certain models, this transparent architecture provides value through its explainable process - particularly important in adversarial settings like legal proceedings.

\section{Conclusion and Future Work}
\label{section:conclusion}

Our structured prompting framework integrating neural and symbolic approaches demonstrated significant effectiveness in hearsay determination tasks, with OpenAI \textit{o}-family models showing substantial improvements - \textit{o1} achieving 0.929 F1 score (compared to 0.714 with few-shot prompting) and \textit{o3-mini} reaching 0.867 (compared to 0.74 with few-shot prompting). These results suggest the potential of our hybrid approach combining LLMs' interpretative capabilities with symbolic verification's logical consistency.

The most significant findings include substantial performance improvements when implementing explicit predicate choices, indicating that complementary predicates may help reduce hallucination and enhance logical consistency. We observed model-dependent performance patterns revealing variable sensitivity to prompting strategies, pointing to the importance of tailoring techniques to specific model architectures, while the structured decomposition approach provided transparent reasoning paths that enhance explainability regardless of performance outcomes.

Future research should explore extending the framework through: (1) incorporating deontic logic to better capture complex legal concepts; (2) developing model-driven task definitions to reduce reliance on manual predicate creation; (3) exploring richer entity-relationship models while maintaining verifiability; and (4) evaluating cross-domain applications to determine whether our findings represent broader principles of neural-symbolic integration. Multi-agent frameworks where agents negotiate task definitions could also better model adversarial reasoning processes in legal argumentation \cite{Choinski2009}.

\bibliographystyle{unsrt}  


\end{document}